%% file: main.tex
\documentclass[11pt]{article}
\usepackage{coling2018}
\usepackage{times}
\usepackage{url}
\usepackage{latexsym}
\usepackage{amsmath}
\usepackage{amsfonts}
\usepackage{float}
\usepackage{graphicx}

\usepackage{multirow}
\usepackage{url}
\usepackage{array,multirow}
\usepackage{amssymb}
\usepackage[vlined,linesnumbered]{algorithm2e}
\usepackage{bm}
\usepackage[draft,hidelinks]{hyperref}

\usepackage{cleveref}  

\usepackage{caption}
\usepackage{subcaption}
\newcommand{\tstart}{\text{\textless}}
\newcommand{\tend}{\text{\textgreater}}

\title{Two Local Models for Neural Constituent Parsing}

\author{Zhiyang Teng \and Yue Zhang \\
        Singapore University of Technology and Design \\
        {\tt zhiyang\_teng@mymail.sutd.edu.sg} \\
        {\tt yue\_zhang@sutd.edu.sg} }

\date{}

\begin{document}
\maketitle
\input{part/abstract2.tex}
\input{part/intro2.tex} 
\input{part/model2.tex}
\input{part/experiment2.tex}
\input{part/analysis2.tex}
\input{part/relatedwork.tex}
\input{part/conclusion.tex}
\input{part/ack.tex}

\bibliographystyle{acl}
\bibliography{tacl}

\end{document}

%% file: part/abstract2.tex
Non-local features have been exploited by syntactic parsers for capturing dependencies between sub output structures.  Such features have been a key to the success of state-of-the-art statistical parsers. With the rise of deep learning, however, it has been shown that local output decisions can give highly competitive accuracies, thanks to the power of dense neural input representations that embody global syntactic information. We investigate two conceptually simple local neural models for constituent parsing, which make local decisions to constituent spans and CFG rules, respectively. Consistent with previous findings along the line, our best model gives highly competitive results, achieving the labeled bracketing F1 scores of 92.4\% on PTB and 87.3\% on CTB 5.1.

%% file: part/intro2.tex
\section{Introduction}
\blfootnote{
    %
    %
    \hspace{-0.65cm}  
    %
    %
    %
    %
     \hspace{-0.65cm}  
     This work is licensed under a Creative Commons 
     Attribution 4.0 International License.
     License details:
     \url{http://creativecommons.org/licenses/by/4.0/}.
}
Non-local features have been shown crucial for statistical parsing \cite{huang:2008:ACLMain,zhangnivre2011transition}. 
For dependency parsing, High-order dynamic programs \cite{koocollins:2010:ACL}, integer linear programming \cite{martins:2010:EMNLP} and dual decomposition \cite{koo:2010:EMNLP} techniques have been exploited by graph-based parser to integrate non-local features. Transition-based parsers \cite{Nivre03anefficient,Nivre:2008:CL,zhangnivre2011transition,bohnet:2010:coling,huangfayongguo:2012:NAACL} are also known for leveraging non-local features for achieving high accuracies. For most state-of-the-art statistical parsers, a global training objective over the entire parse tree has been defined to avoid label bias \cite{lafferty2001conditional}.

For neural parsing, on the other hand, local models have been shown to give highly competitive accuracies \cite{james2016span,sternandreasklein:2017:Long} as compared to those that employ long-range features  \cite{watanabe2015transition,zhou:2015:sentloss,andor:2016:global,durrett2015crf}. Highly local features have been used in recent state-of-the-art models \cite{sternandreasklein:2017:Long,Dozat2016biaffine,shi-huang-lee:2017:EMNLP2017}. In particular, \newcite{Dozat2016biaffine} show that a locally trained arc-factored model can give the best reported accuracies on dependency parsing. The surprising result has been largely attributed to the representation power of long short-term memory (LSTM) encoders \cite{KiperwasserG16a}.

An interesting research question is to what extent the encoding power can be leveraged for constituent parsing. We investigate the problem by building a chart-based model that is local to unlabeled constituent spans \cite{abney1991parsing} and CFG-rules, which have been explored by early PCFG models \cite{collins2003head,klein2003unlex}. In particular, our models first predict unlabeled CFG trees leveraging bi-affine modelling \cite{Dozat2016biaffine}. Then, constituent labels are assigned on unlabeled trees by using a tree-LSTM to encode the syntactic structure, and a LSTM decoder for yielding label sequences on each node, which can include unary rules. 

Experiments show that our conceptually simple models  give highly competitive performances compared with the state-of-the-art. Our best models give labeled bracketing F1 scores of 92.4\% on PTB and 87.3\% on CTB 5.1 test sets, without reranking, ensembling and external parses. We release our code at \url{https://github.com/zeeeyang/two-local-neural-conparsers}.

%% file: part/model2.tex
\section{Model}
\begin{figure*}[t]
\begin{subfigure}[b]{0.5\textwidth}
\centering
   \includegraphics[width=0.6\textwidth]{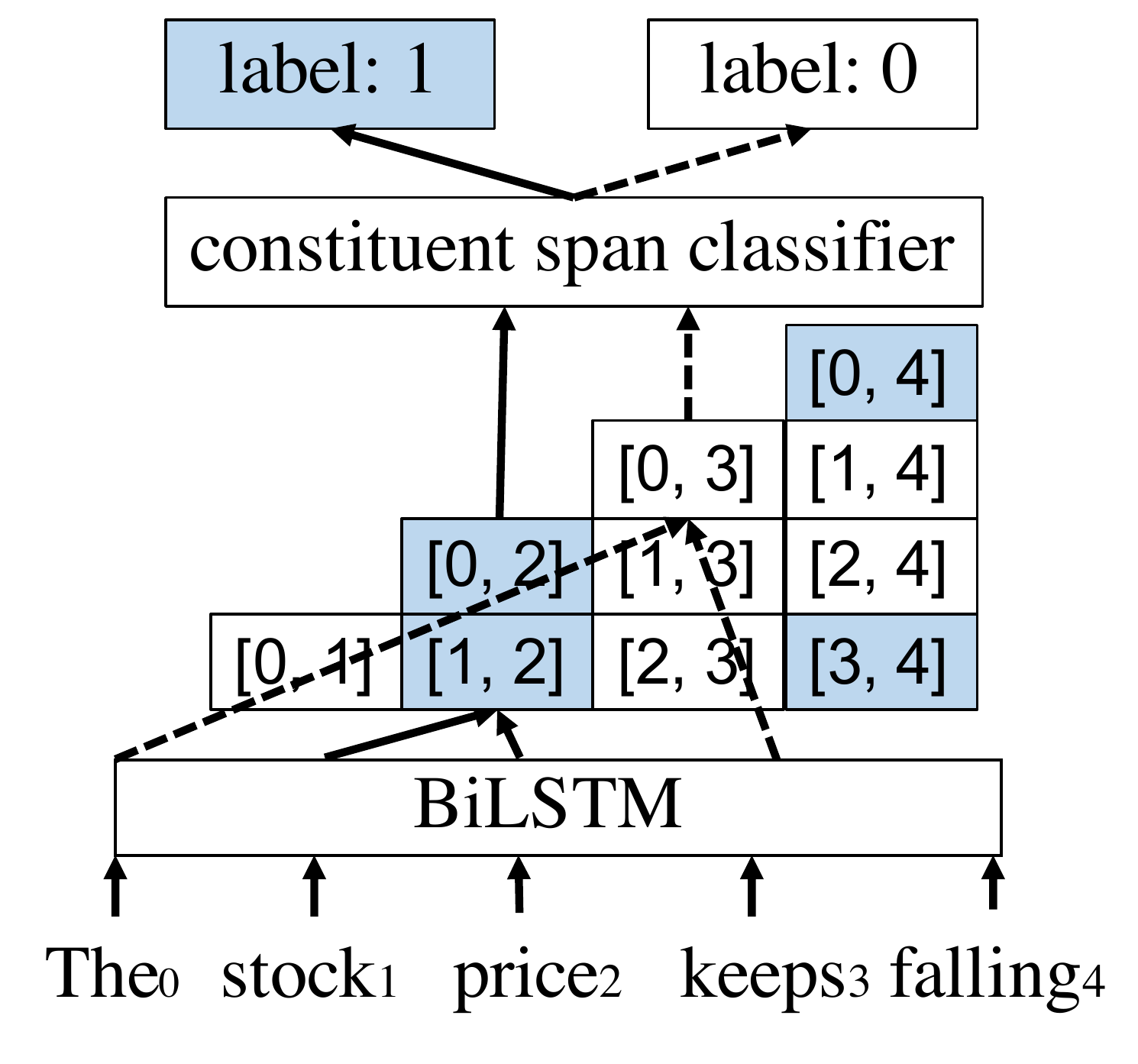}
   \caption{BiLSTM based constituent span classifier.}
   \label{fig:spanclas}
\end{subfigure}
\begin{subfigure}[b]{0.45\textwidth}
\centering
\includegraphics[width=0.8\textwidth]{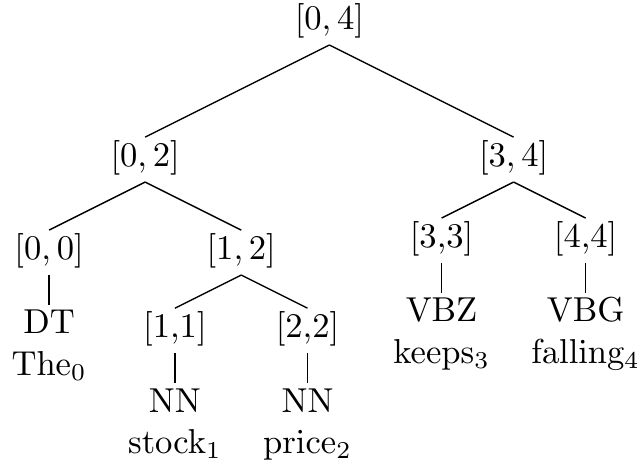}
\caption{unlabeled binarized parse tree.}
\label{fig:binarized}
\end{subfigure}
\begin{subfigure}[b]{0.55\textwidth}
\centering
\includegraphics[width=\textwidth]{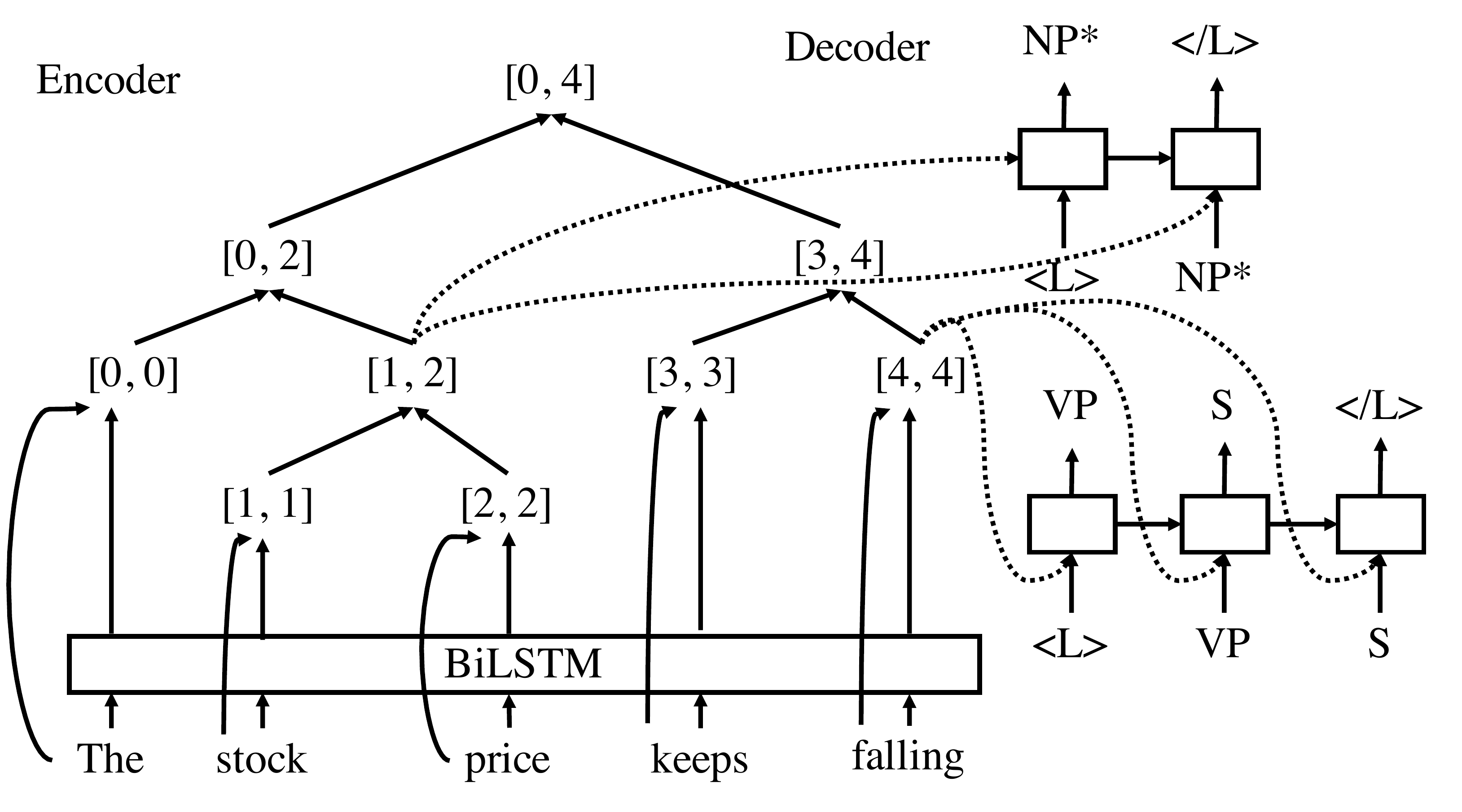}
\caption{Label generator for two example spans. \textbf{NP*} is an intermediate constituent label. }
\label{fig:binlabeltree}
\end{subfigure}
\begin{subfigure}[b]{0.4\textwidth}
    \centering
\includegraphics[width=\textwidth]{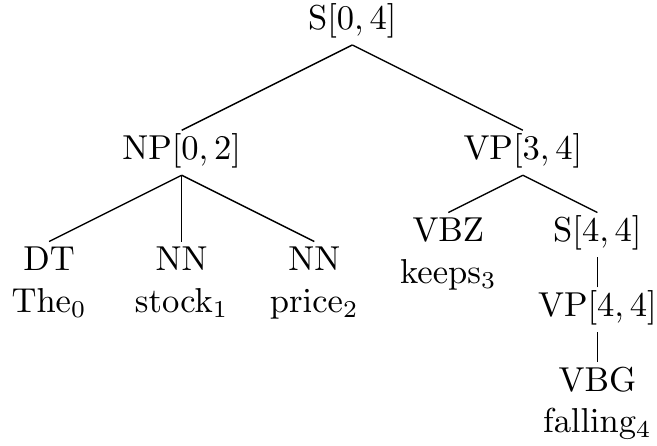}
\caption{Final example parse tree.}
\label{fig:fulltree}
\end{subfigure}
\caption{An example workflow of our parsers for the sentence ``The stock price keeps falling''. We annotate every non-terminal span with its covered span range. Figure \ref{fig:spanclas} shows constituent span classifiers making 0/1 decisions for all possible spans. Based on the local classification probabilities, we obtain an unlabeled binarized parse tree (Figure \ref{fig:binarized}) using binary CKY parsing algorithms. We then hierarchically generate labels for each span (Figure \ref{fig:binlabeltree}) using encoder-decoder models. Figure \ref{fig:fulltree} shows the final output parse tree after debinarization.}
\label{fig:all}
\end{figure*}

Our models consist of an unlabeled binarized tree parser and a label generator. 
Figure \ref{fig:all} shows a running example of our parsing model.  
The unlabeled parser (Figure \ref{fig:spanclas}, \ref{fig:binarized}) learns an unlabeled parse tree using simple BiLSTM encoders \cite{hochreiter1997long}.  The label generator (Figure \ref{fig:binlabeltree}, \ref{fig:fulltree})  predicts constituent labels for each span in the unlabeled tree using tree-LSTM  models. 

In particular, we design two different classification models for unlabeled parsing: the \textbf{span model} (Section \ref{sec:binary}) and the \textbf{rule model} (Section \ref{sec:biaffine}). 
The span model identifies the probability of an arbitrary span being a constituent span. 
For example, the span $[1,2]$ in Figure \ref{fig:spanclas} belongs to the correct parse tree (Figure \ref{fig:fulltree}). Ideally, our model assigns a high probability to this span.  
In contrast, the span $[0,3]$ is not a valid constituent span and our model labels it with 0. 
Different from the span  model,  the rule model  considers the probability  $P([i, j] \rightarrow [i, k] [k+1, j]|S)$ for the  production rule that the span $[i, j]$ is composed by two children spans $[i, k]$ and $[k+1, j]$, where $i\leq k < j$.    For example, in Figure \ref{fig:spanclas}, the rule model assigns high probability to the rule $[0, 2] \rightarrow [0, 0] [1, 2]$ instead of the rule $[0, 2] \rightarrow [0, 1] [2, 2]$.  Given the local probabilities, we use CKY algorithm to find the unlabeled binarized parses. 

The label generator encodes a binarized unlabeled tree and to predict  constituent labels for every span. 
The encoder is a binary tree-LSTM \cite{tai15treelstm,zhu15treelstm},  which recursively composes the representation vectors for tree nodes bottom-up. 
Based on the representation vector of a constituent span, a LSTM decoder \cite{cho2014properties,sutskever2014sequence} generates chains of constituent labels, which can represent unary rules. For example, the decoder outputs ``VP $\rightarrow$S$ \rightarrow$ $<$/L$>$'' for the span [4, 4]  and ``NP$\rightarrow$ $<$/L$>$'' for the span [0,2] in Figure \ref{fig:binlabeltree} where $<$/L$>$ is a stopping symbol. 
\subsection{Span Model}
\label{sec:binary}
 Given an unlabeled binarized tree $T_{ub}$ for the sentence $S$, $S= w_0, w_1 \dots w_{n-1}$,  the span model trains a neural network model $P(Y_{[i,j]}|S, \Theta)$ to distinguish  constituent spans from  non-constituent spans, where $0 \leq i \leq n-2$, $ 1 \leq j < n$, $i < j$. $Y_{[i, j]} =1 $ indicates the span [i, j] is a constituent span ($[i, j] \in T_{ub}$), and  $Y_{[i, j]} =0 $ for otherwise, $\Theta$ are model parameters.  We do not model spans with length 1 since the span $[i, i]$ always belongs to $T_{ub}$. 

\begin{figure*}[!t]
\begin{subfigure}[b]{0.48\textwidth}
   \includegraphics[width=\textwidth]{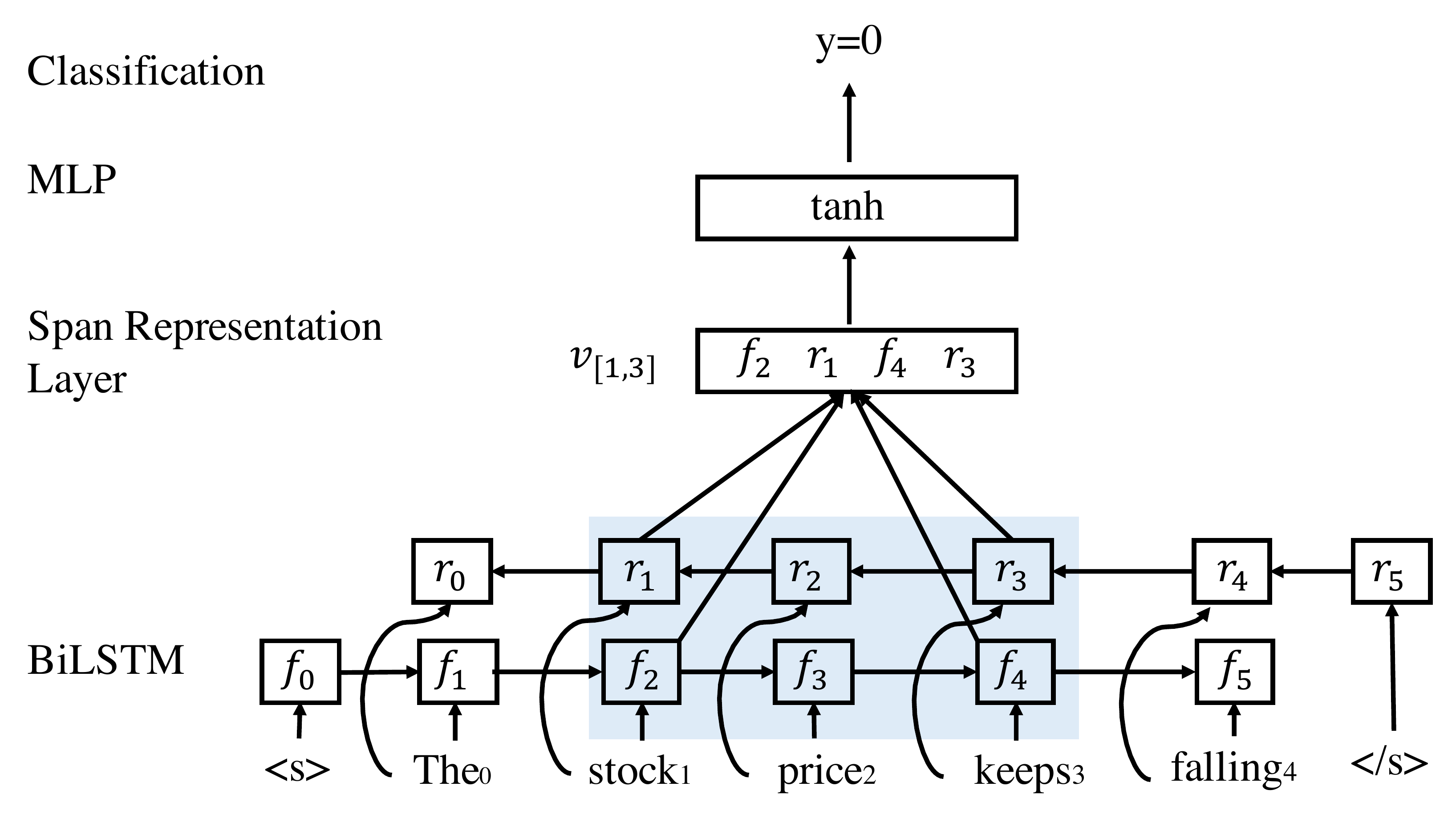}
   \caption{Span model.  }
   \label{fig:binclas}
\end{subfigure}
\begin{subfigure}[b]{0.48\textwidth}
   \includegraphics[width=\textwidth]{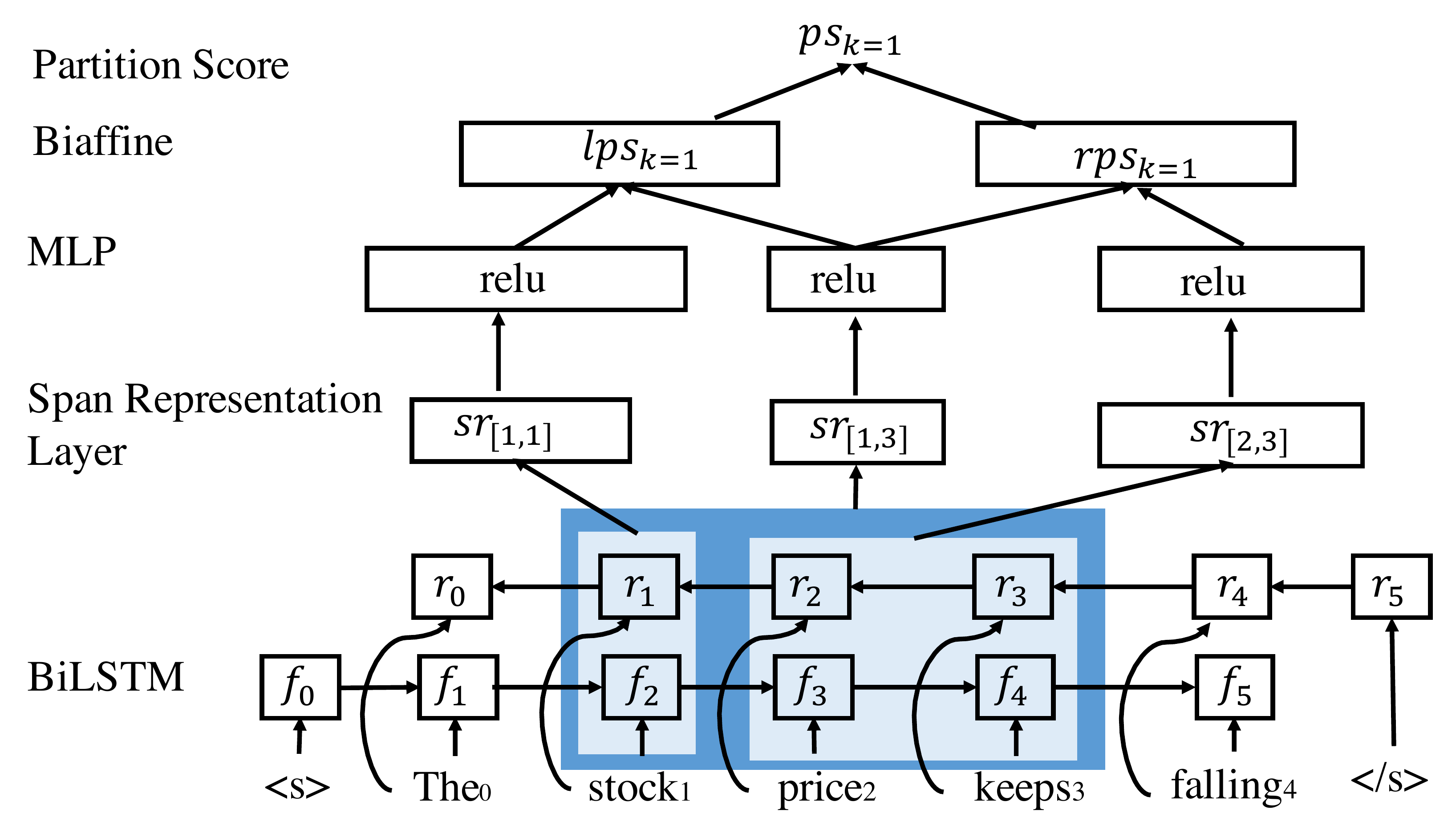}
   \caption{Rule model.   }
   \label{fig:biaffine}
\end{subfigure}
\caption{Neural network structures for span and rule models using BiLSTM encoders.  }
\label{fig:model}
\end{figure*}

\textbf{Network Structure.} Figure \ref{fig:binclas} shows the neural network structures for the binary classification model. In the bottom, a bidirectional LSTM layer encodes the input sentence to extract non-local features. In particular, we append a starting symbol $<$s$>$ and an ending symbol $<$/s$>$ to the left-to-right LSTM and the right-to-left LSTM, respectively.   We denote the output hidden vectors of the left-to-right LSTM and the right-to-left LSTM  for $w_0, w_1, \dots, w_{n-1}$ 
is $\mathbf{f}_1, \mathbf{f}_2, \dots, \mathbf{f}_n$ and $\mathbf{r}_0, \mathbf{r}_1, \dots, \mathbf{r}_{n-1}$, respectively.  
We obtain the representation vector $\mathbf{v}[i, j]$ of the span $[i, j]$ by simply concatenating the bidirectional output vectors at the input word $i$ and the input word $j$,
\begin{equation}
\mathbf{v}[i,j]=[\mathbf{f}_{i+1};\mathbf{r}_{i}; \mathbf{f}_{j+1}; \mathbf{r}_{j}].
\label{eq:v}
\end{equation}
$\mathbf{v}[i, j]$ is then passed through a nonlinear transformation layer and  the probability distribution 
$P(Y_{[i,j]}|S, \Theta)$ is given by
\begin{equation}
\label{eq:binaryp}
        \mathbf{o}[i,j] = \tanh (\mathbf{W}_o \mathbf{v}[i,j] + \mathbf{b}_o),\ \  
        \mathbf{u}[i,j] = \mathbf{W}_u \mathbf{o}[i,j] + \mathbf{b}_u, \ \ 
        P(Y_{[i,j]}|S, \Theta) = \text{softmax}(\mathbf{u}[i,j]), 
\end{equation}
where $\mathbf{W}_o, \mathbf{b}_o, \mathbf{W}_u$ and $\mathbf{b}_u$ are model parameters. 

\textbf{Input Representation.}  Words and part-of-speech (POS) tags are integrated  to obtain the input representation vectors.  Given a word $w$, its corresponding characters $c_0,\dots, c_{|w|-1}$ and POS tag $t$, first, we obtain the word embedding $\mathbf{E}_{word}^w$, character embeddings $\mathbf{E}_{char}^{c_0},\dots, \mathbf{E}_{char}^{c_{|w|-1}}$, and POS tag embedding $\mathbf{E}_{pos}^t$ using lookup operations. Then a bidirectional LSTM is used to extract character-level features. Suppose that the last output vectors of the left-to-right and right-to-left LSTMs are $\mathbf{h}_{char}^f$ and $\mathbf{h}_{char}^r$, respectively.  The final input vector $\mathbf{x}_{input}$ is given by 
\begin{equation}
        \mathbf{x}_{char} = \tanh (\mathbf{W}_{char}^l \mathbf{h}_{char}^l +  \mathbf{W}_{char}^r \mathbf{h}_{char}^r + \mathbf{b}_{char} ),  \ \ 
        \mathbf{x}_{input} = [ \mathbf{E}_{word}^w + \mathbf{x}_{char};  \mathbf{E}_{pos}^t],
\label{eq:xinput}
\end{equation}
where $\mathbf{W}_{char}^l$, $\mathbf{W}_{char}^r$ and $\mathbf{b}_{char}$ are model parameters. 

\textbf{Training objective.} The training objective is to maximize the probabilities of $P(Y_{[i,j]}=1|S, \Theta)$ for spans $[i,j] \in T_{ub}$ and minimize the probabilities of $P(Y_{[i,j]}=1|S, \Theta)$ for spans $[i,j] \notin T_{ub}$ at the same time. 
Formally, the training loss for binary span classification ${\mathcal{L}}_{\text{binary}}$ is given by
\begin{equation}
\begin{split}
   {\mathcal{L}}_{\text{binary}} = -\sum_{[i,j] \in T_{ub}}&{ \log P(Y_{[i,j]} = 1 | S, \Theta)}  -\sum_{[i,j] \notin T_{ub}}{ \log P(Y_{[i,j]} = 0 |S, \Theta)}, \\ 
    &(0 \leq i \leq n-2, 1 \leq j < n, i < j)
\end{split}
\label{eq:binloss}
\end{equation}
For a sentence with length $n$, there are $\frac{ n (n -1) } {2}$ terms in total in Eq \ref{eq:binloss}.

\textbf{Neural CKY algorithm.} The unlabeled production probability for the rule $r:[i, j]\rightarrow[i, k][k+1, j]$  given by the binary classification model is,  
\begin{equation*}
P(r | S, \Theta) = P( Y_{[i, k]} = 1 |S, \Theta )P( Y_{[k+1, j]} = 1 |S, \Theta).
\end{equation*}
 During decoding, we find the optimal parse tree $T_{ub}^{*}$ using the CKY algorithm. Note that our CKY algorithm is different from the standard CKY algorithm mainly in that there is no explicit phrase rule probabilities being involved. Hence our model can be regarded as a zero-order constituent tree model, which is the most {\it local}. All structural relations in a constituent tree must be implicitly captured by the BiLSTM encoder over the sentence alone. 

\textbf{Multi-class Span Classification Model.} The previous model preforms binary classifications to identify constituent spans. In this way,  the classification model  only captures the existence of constituent labels but does not leverage constituent label type information. In order to incorporate the syntactic label information into the span model, we use a multi-class classification model $P(Y_{[i,j]}=c|S, \Theta)$ to describe the probability that $c$ is a constituent label for span $[i,j]$.  The network structure is the same as the binary span classification model except the last layer.  For the last layer, given $\mathbf{o}_{[i,j]}$ in Eq \ref{eq:binaryp},  $P(Y_{[i,j]}=c|S, \Theta)$ is calculated by,
\begin{equation}
\label{eq:multip}
        \mathbf{m}[i,j] = \mathbf{W}_m \mathbf{o}[i,j] + \mathbf{b}_m, \
        P(Y_{[i,j]}=c|S, \Theta) = \text{softmax}(\mathbf{m}[i,j])_{[c]}.
\end{equation}
Here $\mathbf{W}_m, \mathbf{b}_m, \mathbf{W}_m$ and $\mathbf{b}_m$ are model parameters. The subscript $[c]$ is to pick the probability for the label $c$. The training loss is,
\begin{equation}
\begin{split}
    \mathcal{L}_{\text{multi}} = -\sum_{ [i,j] \in T_{ub} } \sum_{c \in \textsc{Gen}[i,j], c\neq \tstart/L\tend }&{ \log P(Y_{[i,j]} = c | S, \Theta)} 
    -\sum_{[i,j] \notin T_{ub}}{ \log P(Y_{[i,j]} = \tstart /L\tend  |S, \Theta)}, \\
    & (0 \leq i \leq n-2, 1 \leq j < n, i < j)\\
\end{split}
\label{eq:mutiloss}
\end{equation}
Note that there is an additional sum inside the first term in Eq \ref{eq:mutiloss}, which is different from the first term in Eq \ref{eq:binloss}. $\textsc{Gen}[i,j]$ denotes the label set of span $[i,j]$. This is to say that we treat all constituent labels equally of a unary chain. For example, suppose there is a unary chain S$\rightarrow$VP in span [4,4]. For this span, we hypothesize that both labels are plausible answers and pay equal attentions to VP and S during training. For the second term in Eq \ref{eq:mutiloss}, we maximize the probability of the ending label for non-constituent spans. 

For decoding, we transform the multi-class probability distribution into a binary probability distribution by using,
\begin{equation*}
P(Y_{[i,j]}=1|S, \Theta) = \sum_{c, c\neq \tstart/L\tend }{ P(Y_{[i,j]} = c | S, \Theta)},    \ \ 
     P(Y_{[i,j]}=0|S, \Theta) = P(Y_{[i,j]}=\tstart/L\tend |S, \Theta)
\label{eq:transform}
\end{equation*}
In this way, the probability of a span being a constituent span takes all possible syntactic labels into considerations. 

\subsection{Rule Model}
\label{sec:biaffine}

The rule model directly calculates the probabilities of all possible splitting points $k$ ($ i\leq k < j)$ for the span $[i, j]$. Suppose the partition score of splitting point $k$ is $ps_k$. The unlabeled production probability for the rule $r:[i, j]\rightarrow[i, k][k+1, j]$ is given by a softmax distribution, 
\begin{equation*}
 P([i, j] \rightarrow [i, k] [k+1, j]|S, \Theta) =  \frac{\exp(ps_k)} { \sum_{k^\prime=i}^{j-1}{ \exp(ps_{k^{\prime}})}}. 
\end{equation*}
The training objective is to minimize the log probability loss of all unlabeled production rules.
\begin{equation*}
{\mathcal{L}}_{\text{rule}} =  -\sum_{r \in {T_{ub}}} \log P(r: [i, j] \rightarrow [i, k] [k+1, j]|S, \Theta)
\end{equation*}

The decoding algorithm is the standard CKY algorithm, which we omit here. The rule model can be regarded as a first-order constituent model, with the probability of each phrase rule being modeled. However, unlike structured learning algorithms \cite{finkel2008crf,carreras2008tag}, which use a global score for each tree, our model learns each production rule probability individually. Such local learning has traditionally been found subjective to label bias \cite{lafferty2001conditional}. Our model relies on input representations solely for resolving this issue.

\textbf{Span Representation.} Figure \ref{fig:biaffine} shows one possible network architecture for the rule model by taking the partition point $k=1$ for the span $[1, 3]$ as an example. 
The BiLSTM encoder layer in the bottom is the same as that of  the previous span classification model. We obtain the span representation vectors using difference vectors \cite{wang2016graph,james2016span}. Formally, the span representation vector $\mathbf{sr}[i, j]$ is given by,
\begin{equation}
\begin{split}
\mathbf{s}[i, j]&=[\mathbf{f}_{j+1} - \mathbf{f}_{i}; \mathbf{r}_{i} - \mathbf{r}_{j+1}],  \\ 
\mathbf{sr}[i, j]&=[ \mathbf{s}[0, i-1]; \mathbf{s}[i, j]; \mathbf{s}[j+1, n-1]]. 
    \label{eq:spanrep}
\end{split}
\end{equation}
We first combine the difference vectors $(\mathbf{f}_{j+1} - \mathbf{f}_{i})$ and $(\mathbf{r}_{i} - \mathbf{r}_{j+1})$ to obtain a simple span representation vector $\mathbf{s}[i, j]$.  
In order to take more contextual information such as $\mathbf{f}_p$ where $p>j+1$ and $\mathbf{r}_q$ where $q<i$, we concatenate $\mathbf{s}[0, i-1]$,  $\mathbf{s}[i, j]$, and $\mathbf{s}[j+1, n-1]$ to produce the final span representation vector $\mathbf{sr}[i, j]$.  We then transform $\mathbf{sr}[i,j]$ to an output vector $\mathbf{r}[i,j]$ using an activation function $\phi$,
\begin{equation}
\label{eq:r}
         \mathbf{r}[i,j] = \phi (\mathbf{W}_r^{M} \mathbf{sr}[i,j] + \mathbf{b}_r^{M}), 
\end{equation}
where $\mathbf{W}_r^{M}$ and $\mathbf{b}_r^{M}$ and model parameters, and $M$ is a parameter set index. We use separate parameters for the nonlinear transforming layer. $M\in \{P, L, R\}$ are for the parent span $[i, j]$, the left child span $[i,k]$ and the right child span $[k+1, j]$, respectively.  

After obtaining the span representation vectors, we use these vectors to calculate the partition score $ps_k$. In particular, we investigate two scoring methods.  

\textbf{Linear Model.} In the linear model, the partition score is calculated by a linear affine transformation. For the splitting point $k$, 
\begin{equation*}
    ps_k = \mathbf{w}_{ll, k}^T \mathbf{r}[i, k] + \mathbf{w}_{lr, k}^T \mathbf{r}[k+1, j] + {b}_{ll,k}
\end{equation*}
where $\mathbf{w}_{ll, k}^T$ and $\mathbf{w}_{ll, k}^T$ are two vectors, and ${b}_{ll,k}$ is a size 1 parameter. 

\textbf{Biaffine model.} Since the possible splitting points for spans are varied with the length of span, we also try a biaffine scoring model (as shown in Figure \ref{fig:biaffine}), which is good at handling variable-sized classification problems \cite{Dozat2016biaffine,ma2017neural}. The biaffine model  produces the score ${lps}_k$ between the parent span $[i,j]$ and the left child span $[i, k]$ using a biaffine scorer
\begin{equation}
         {lps}_k = (\mathbf{r}[i,j] \oplus 1)^{T} \mathbf{W}_{pl} (\mathbf{r}[i,k] \oplus 1) 
\end{equation}
where $\mathbf{W}_{pl}$ is model parameters and $\oplus$ denotes vector concatenation. 
Similarly, we calculate the score ${rps}_k$ between the parent span $[i,j]$ and the right child span $[k+1, j]$ using $\mathbf{W}_{pr}$ and $\mathbf{b}_{pr}$ as parameters.  The overall partition score ${ps}_k$ is therefore given by 
\[
{ps}_k  = {lps}_k + {rps}_k.
\]
\subsection{Label Generator}
\label{sec:encdec}
\textbf{Lexicalized Tree-LSTM Encoder.} 
Shown in Figure \ref{fig:binlabeltree}, we use lexicalized tree LSTM \cite{zhiyang2016bitree} for encoding, which shows good representation abilities for unlabeled trees. 
The encoder first propagates lexical information from two children spans to their parent using a lexical gate, then it produces the representation vectors of the parent span by composing the vectors of children spans using a binarized tree-LSTM \cite{tai15treelstm,zhu15treelstm}. 
Formally, the lexical vector $\mathbf{tx}[i,j]$ for the span $[i,j]$ with the partition point at $k$ is defined by:
\begin{equation*}
    \begin{split}
        \mathbf{i}_{[i,j]}^{lex} &= \sigma ( \mathbf{W}_{l}^{lex} \mathbf{tx}{[i, k]} + \mathbf{W}_{r}^{lex} \mathbf{tx}{[k+1, j]} +    \mathbf{W}_{lh}^{lex}\mathbf{h}_{[i,k]} + \mathbf{W}_{rh}^{lex}\mathbf{h}_{[k+1,j]} +  \mathbf{b}_{lex})\\
        \mathbf{tx}[i,j] &= \mathbf{i}_{[i,j]}^{lex} \odot \mathbf{tx}{[i, k]} + (\mathbf{1.0} - \mathbf{i}_{[i,j]}^{lex}) \odot \mathbf{tx}{[k+1, j]}, \\
    \end{split}
\end{equation*}
where $\mathbf{W}_{l}^{lex}$, $\mathbf{W}_{r}^{lex}$ and $\mathbf{b}_{lex}$ are model parameters, $\odot$ is element-wise multiplication and $\sigma$ is the logistic function.  The lexical vector $\mathbf{tx}[i,i]$ for the leaf node $i$ is the concatenate of the output vectors of the BiLSTM encoder and the input representation $\mathbf{x}_{\textbf{input}}[i]$ (Eq \ref{eq:xinput}), as shown in Figure \ref{fig:binlabeltree}.    

The output state vector $\mathbf{h}[i,j]$ of the span $[i,j]$  given by a binary tree LSTM encoder is, 
\begin{equation*}
    \begin{split}
        &\mathbf{i}_{p} = \sigma (   \mathbf{W}_{1} \mathbf{tx}_{p} +
         \mathbf{W}_{2} \mathbf{h}_{l} + \mathbf{W}_{3} \mathbf{c}_{l} + 
          \mathbf{W}_{4} \mathbf{h}_{r}  + \mathbf{W}_{5} \mathbf{c}_{r} + \mathbf{b}_1 ), \\ 
          &\mathbf{f}_{p}^{l} = \sigma (\mathbf{W}_{6} \mathbf{tx}_{p} +
         \mathbf{W}_{7} \mathbf{h}_{l} + \mathbf{W}_{8} \mathbf{c}_{l}  + 
          \mathbf{W}_{9} \mathbf{h}_{r} + \mathbf{W}_{10} \mathbf{c}_{r} + \mathbf{b}_2 ),  \\
         &\mathbf{f}_{p}^{r} = \sigma (   \mathbf{W}_{11} \mathbf{tx}_{p} +
         \mathbf{W}_{12} \mathbf{h}_{r} + \mathbf{W}_{13} \mathbf{c}_{l} +
         \mathbf{W}_{14} \mathbf{h}_{r} + \mathbf{W}_{15} \mathbf{c}_{r} + \mathbf{b}_{3} ), \\ 
        &\mathbf{g}_{p} =\tanh ( \mathbf{W}_{16} \mathbf{tx}_{p} + 
         \mathbf{W}_{17} \mathbf{h}_{l} + 
          \mathbf{W}_{18} \mathbf{h}_{r}   + \mathbf{b}_{4} ),  \\
         &\mathbf{c}_{p}  = \mathbf{f}_{p}^{l}  \odot  \mathbf{c}_{l}  +\mathbf{f}_{p}^{r}  \odot  \mathbf{c}_{r}+ \mathbf{i}_{p} \odot  \mathbf{g}_{p}, \\ 
         &\mathbf{o}_{p} = \sigma (   \mathbf{W}_{19} \mathbf{tx}_{p} +
         \mathbf{W}_{20} \mathbf{h}_{p} + 
         \mathbf{W}_{21}  \mathbf{h}_{r}   +
          \mathbf{W}_{22} \mathbf{c}_{p}  + \mathbf{b}_5),  \ \ 
         \mathbf{h}_{p} =  \mathbf{o}_{p} \odot \tanh(\mathbf{c}_{p}).\\
    \end{split}
\end{equation*}
Here the subscripts $p$, $l$ and $r$ denote $[i, j]$, $[i, k]$ and $[k+1,j]$, respectively. 

\textbf{Label Decoder.}  Suppose that the constituent label chain for the span $[i,j]$ is ($\text{YL}_{[i,j]}^0, \text{YL}_{[i,j]}^1, \dots, \text{YL}_{[i,j]}^m$). The decoder for the span $[i,j]$ learns a conditional language model depending on the output vector $\mathbf{h}[i,j]$ from the tree LSTM encoder. Formally, the probability distribution of generating the label at time step $z$ is given by, 
\begin{equation*}
    P(\text{YL}_{[i,j]}^{z} | T_{ub}, \text{YL}_{[i,j]}^{z< m}) = 
     \text{softmax}\Big(g(\mathbf{h}[i,j], \mathbf{E}_{label} (\text{YL}_{[i,j]}^{z-1}), \mathbf{d}_{z-1})\Big), 
\end{equation*}
where $\text{YL}_{[i,j]}^{z< m}$ is the decoding prefix, $\mathbf{d}_{z-1}$ is the state vector of the decoder LSTM and $\mathbf{E}_{label} (\text{YL}_{[i,j]}^{z-1})$ is the embedding of the previous output label.  

The training objective is to minimize the negative log-likelihood of the label generation distribution, 
\begin{equation*}
\begin{split}
        &\mathcal{L}_{\text{label}}{[i,j]} =  -\sum_{z=0}^{m} \log P(\text{YL}_{[i,j]}^{z} | T_{ub}, \text{YL}_{[i,j]}^{z< m}), \\
         &\mathcal{L}_{\text{label}} = \sum_{[i,j] \in T_{ub} }  \mathcal{L}_{\text{label}}{[i,j]}. 
         \end{split}
\end{equation*}
\subsection{Joint training}
In conclusion, each model contains an unlabeled structure predictor and a label generator.  The latter is the
same for all models. All the span models perform binary classification. The difference is that 
BinarySpan doesn't consider label information for unlabeled tree prediction. While MultiSpan guides 
unlabeled tree prediction with such information, simulating binary classifications. The unlabeled parser and the label generator share parts of the network components, such as word embeddings, char embeddings, POS embeddings and the BiLSTM encoding layer. 
We jointly train the unlabeled parser and the label generator for each model by minimizing the overall loss 
\[
     \mathcal{L}_{\text{total}} =  \mathcal{L}_{\text{parser}} +  \mathcal{L}_{\text{label}}  + \frac{\lambda}{2} ||\Theta||^2,
\]
where $\lambda$ is a regularization hyper-parameter. We set $ \mathcal{L}_{\text{parser}}= \mathcal{L}_{\text{binary}}$ or $ \mathcal{L}_{\text{parser}}= \mathcal{L}_{\text{multi}}$ and $ \mathcal{L}_{\text{parser}}= \mathcal{L}_{\text{rule}}$ when using the binary span classification model, the multi-class model and the rule model, respectively. 

%% file: part/experiment2.tex
\section{Experiments}

\subsection{Experimental Settings}

\textbf{Data.} We perform  experiments  for both English and Chinese. 
Following standard conventions, our English data are obtained from the Wall Street Journal (WSJ) of the Penn Treebank (PTB) \cite{marcus1993building}. Sections 02-21, section 22 and section 23 are used for training, development and test sets, respectively. 
Our Chinese data are the version 5.1 of the Penn Chinese Treebank (CTB) \cite{xue2005penn}. The training set consists of articles 001-270 and 440-1151, the development set contains articles 301-325 and the test set includes articles 271-300. We use automatically reassigned POS tags in the same way as \newcite{james2016span} for English and \newcite{dyer2016rnng} for Chinese. 

We use ZPar \cite{zhang2011syntactic}\footnote{https://github.com/SUTDNLP/ZPar} to binarize both English and Chinese data with the  head rules of \newcite{collins2003head}.  The head directions of the binarization results are ignored during training. The types of English and Chinese constituent span labels after binarization are 52 and 56, respectively. The maximum number of greedy decoding steps for generating consecutive constituent labels is limited to 4 for both English and Chinese. We evaluate parsing performance in terms of both unlabeled bracketing metrics and labeled bracketing metrics including unlabeled F1 (UF)\footnote{For UF, we exclude the sentence span [0,n-1] and all spans with length 1.}, labeled precision (LP), labeled recall (LR) and labeled bracketing F1 (LF) after debinarization using EVALB\footnote{http://nlp.cs.nyu.edu/evalb}.  

\textbf{Unknown words.} For English, we combine the methods of \newcite{dyer2016rnng}, \newcite{KiperwasserTACLeasyfirst} and  \newcite{james2016span} to handle unknown words. In particular, we first map all words (not just singleton words) in the training corpus into unknown word classes using the same rule as \newcite{dyer2016rnng}.  During each training epoch, every word $w$ in the training corpus is stochastically mapped into its corresponding unknown word class $unk_w$ with probability $P( w \rightarrow unk_w) = \frac{ \gamma } {  \gamma + \#w }$, where $\#w$ is the frequency count and $\gamma$ is a control parameter. 
Intuitively, the more times a word appears, the less opportunity it will be mapped into its unknown word type. There are 54  unknown word types for English.  Following \newcite{james2016span}, $\gamma=0.8375$.  
For Chinese, we simply use one unknown word type to dynamically replace singletons words with a probability of 0.5.  

\begin{table}[!t]
    \centering
    \small
    \setlength{\tabcolsep}{3.5pt}
  \begin{tabular}{l|c|l|c}
    \hline
    \bf hyper-parameter & \bf value & \bf hyper-parameter & \bf value\\
    \hline
    Word embeddings & English: 100 Chinese: 80 &
    Word LSTM layers & 2  \\
    Word LSTM hidden units & 200 &
    Character embeddings & 20 \\
    Character LSTM layers & 1 &
    Character LSTM hidden units & 25 \\
    Tree-LSTM hidden units & 200 &
    POS tag embeddings & 32 \\
    Constituent label embeddings & 32 & 
    Label LSTM layers & 1 \\
    Label LSTM hidden units & 200 & 
    Last output layer hidden units & 128 \\
    Maximum training epochs & 50 &
    Dropout & English: 0.5, Chinese 0.3 \\
    Trainer & SGD &
    Initial learning rate & 0.1 \\
    Per-epoch decay & 0.05 &
    $\phi$ & $\textsc{ELU}$ \cite{elu2015} \\
    \hline
   \end{tabular}
   \caption{Hyper-parameters for training.}
  \label{tab:hyper}
\end{table}

 \textbf{Hyper-parameters.}  Table \ref{tab:hyper} shows all hyper-parameters. These values are tuned using the corresponding development sets. We optimize our models with  stochastic gradient descent (SGD).  The initial learning rate is 0.1. Our model are  initialized with pretrained word embeddings both for English and Chinese. The pretrained word embeddings are the same as those used in \newcite{dyer2016rnng}. The other parameters are initialized according to the default settings of DyNet \cite{neubig2017dynet}. We apply dropout \cite{srivastava2014dropout} to the inputs of every LSTM layer, including the word LSTM layers, the character LSTM layers,  the tree-structured LSTM layers and the constituent label LSTM layers. For Chinese, we find that 0.3 is a good choice for the dropout probability. The number of training epochs is decided by the evaluation performances on development set. In particular, we perform evaluations on development set for every 10,000 examples.  The training procedure stops when the results of next 20 evaluations do not become better than the previous best record.

\subsection{Development Results}
We study the two span representation methods, namely the simple concatenating representation $\mathbf{v}[i,j]$ (Eq \ref{eq:v}) and the  combining of three difference vectors $\mathbf{sr}[i,j]$ (Eq \ref{eq:spanrep}), and the  two representative models, i.e, the binary span classification model (\textbf{BinarySpan}) and the biaffine rule model (\textbf{BiaffineRule}). We investigate appropriate representations for different models on the English dev  dataset. Table \ref{tab:repcompare} shows the effects of  different span representation methods, where $\mathbf{v}[i,j]$ is better for \texttt{BinarySpan} and $\mathbf{sr}[i,j]$ is better for \texttt{BiaffineRule}. When using $\mathbf{sr}[i,j]$ for \texttt{BinarySpan}, the performance drops greatly ($92.17\rightarrow 91.80$). Similar observations can be found when replacing $\mathbf{sr}[i,j]$ with $\mathbf{v}[i,j]$ for \texttt{BiaffineRule}. Therefore, we use $\mathbf{v}[i,j]$ for the span models and $\mathbf{sr}[i,j]$ for the rule  models in latter experiments.  

Table \ref{tab:maindev} shows the main results on the English and Chinese dev sets. For English, \texttt{BinarySpan} acheives 92.17 LF score. The multi-class span classifier (\textbf{MultiSpan}) is much better than \texttt{BinarySpan} due to the awareness of label information. Similar phenomenon can be observed on the Chinese dataset. We also test the  linear rule (\textbf{LinearRule}) methods. For English, \texttt{LinearRule} obtains 92.03 LF score, which is much worse than \texttt{BiaffineRule}.  
In general, the performances of \texttt{BiaffineRule} and \texttt{MultiSpan} are quite close both for English and Chinese.

For \texttt{MultiSpan}, both the first stage (unlabeled tree prediction) and the second stage (label generation) exploit constituent types. We design three development experiments to answer what the accuracy would be like of the predicted labels of the first stage were directly used in the second stage. The first  one doesn't include the label probabilities of the first stage for the second stage.   
For the second experiment, we directly use the  model output from the first setting for decoding,  summing up the label 
classification probabilities of the first stage and the label generation probabilities of the second stage in 
order to make label decisions.  For the third setting, we do the sum-up of label probabilities for the second stage both during training
and decoding. These settings give LF scores of 92.44, 92.49 and 92.44, respectively, which are very 
similar. We choose the first one due to its simplicity. 

\begin{table*}[!t]
\small
\setlength{\tabcolsep}{3.0pt}
\begin{minipage}{.5\linewidth}  
\centering
\begin{tabular}{l|l|c|c|c}
    \hline 
        Model & SpanVec & LP & LR & LF \\
        \hline 
        \multirow{ 2}{*}{BinarySpan} & $\mathbf{v}[i,j]$ & \bf 92.16 & \bf 92.19 & \bf 92.17 \\
          & $\mathbf{sr}[i,j]$ & 91.90	& 91.70 & 91.80 \\
        \hline 
       \multirow{ 2}{*}{BiaffineRule}  & $\mathbf{v}[i,j]$ & 91.79 &	91.67 &	91.73 \\
         & $\mathbf{sr}[i,j]$ & \bf 92.49 &	\bf 92.23 &	\bf 92.36 \\
        \hline
    \end{tabular} 
  \caption{Span representation methods. }
    \label{tab:repcompare} 
\end{minipage}
\begin{minipage}{.5\linewidth}
\centering
\begin{tabular}{l|c|c|c|c|c|c}
\hline
\multirow{2}{*}{Model} & \multicolumn{3}{c|}{English} & \multicolumn{3}{c}{Chinese} \\
\cline{2-7}
      & LP & LR & LF & LP & LR & LF\\
     \hline
     BinarySpan & 92.16 &	92.19 &	92.17 & 91.31 & 90.48 & 90.89\\
     MultiSpan & 92.47	& \bf 92.41	& \bf 92.44 & \bf 91.69	& 90.91	& \bf 91.30\\
     \hline 
     LinearRule & 92.03 &	92.03 & 92.03 & 91.03 & 89.19 &	90.10\\
     BiaffineRule & \bf 92.49 &	92.23 & 92.36& 91.31 &	\bf 91.28	& 91.29\\
\hline 
\end{tabular}
\caption{Main development results.}
\label{tab:maindev}
\end{minipage}
\end{table*}
\subsection{Main Results}

\textbf{English.} Table \ref{tab:expeng} summarizes the performances of various constituent parsers on PTB test set.  \texttt{BinarySpan} achieves 92.1 LF score, outperforming the neural CKY parsing models \cite{durrett2015crf} and the top-down neural parser \cite{sternandreasklein:2017:Long}.  \texttt{MultiSpan} and \texttt{BiaffineRule} obtain similar performances. Both are better than \texttt{BianrySpan}. \texttt{MultiSpan} obtains 92.4 LF score, which is very close to the state-of-the-art result when no external parses are included. An interesting observation is that the model of \newcite{sternandreasklein:2017:Long} show higher LP score than our models (93.2 v.s 92.6), while our model gives better LR scores (90.4 v.s. 93.2). This potentially suggests that the global constraints such as structured label loss used in \cite{sternandreasklein:2017:Long} helps make careful decisions. Our local models are likely to gain a better balance between bold guesses and accurate scoring of constituent spans. Table \ref{tab:unlabeledf1} shows the unlabeled parsing accuracies on PTB test set. \texttt{MultiSpan} performs the best, showing 92.50 UF score. When the unlabeled parser is 100\% correct, \texttt{BiaffineRule} are better than the other two, producing an oracle LF score of 97.12\%, which shows the robustness of our label generator. The decoding speeds of \texttt{BinarySpan} and \texttt{MutliSpan} are similar, reaching about 21 sentences per second. \texttt{BiaffineRule} is much slower than the span models. 
\begin{table}[!t]
  \centering
  \small
  \setlength{\tabcolsep}{3.5pt}
  \begin{tabular}{l|l|l|l|l|l|l|l}
    \hline
    Parser       & LR & LP &  LF & Parser       & LR & LP &  LF\\
    \hline
    \newcite{zhu2013acl} (S) & 91.1 & 91.5 & 91.3 & 
    \newcite{charniak2000maximum}&  89.5 & 89.9 & 89.5 \\
    
    \newcite{mcclosky2006rerank} (S) & 92.1 & 92.5 & 92.3 & 
    \newcite{collins2003head} & 88.1 & 88.3 & 88.2 \\
     
    \newcite{choe2016parsinglm} (S,R,E) & & & 93.8 &
    
    \newcite{sagae2006combine} & 87.8 & 88.1 & 87.9 \\
    
    \newcite{durrett2015crf} (S) & & & 91.1 & 
    \newcite{petrov2007unlex} & 90.1 & 90.2 & 90.1 \\
    
    \newcite{vinyals2015grammar} (S, E) & & & 92.8 &
     \newcite{carreras2008tag} & 90.7 & 91.4 & 91.1 \\
     
    \newcite{charniak2005rerank} (S, R) & 91.2 & 91.8 & 91.5 &
    \newcite{zhu2013acl} & 90.2 & 90.7 & 90.4 \\
    
    \newcite{huang2008forest} (R) & & & 91.7 &
    \newcite{watanabe2015transition} & & & 90.7\\
    
    \newcite{huang2009selftraining} (ST) & 91.1 & 91.6 & 91.3 &
     \newcite{fernandezgonzalez-martins:2015:ACL-IJCNLP} & 89.9 & 90.4 & 90.2 \\
    
    \newcite{huang2010product} (ST) & 92.7 & 92.2 & 92.5 &
   \newcite{james2016span}& 90.5 & 92.1 & 91.3 \\ 
    
    \newcite{shindo2012refined} (E) & & & 92.4 & 
    \newcite{kuncoro2017rnng}  & & & 91.2   \\
    \newcite{socher2013parsing} (R) & & & 90.4 &
    \newcite{liu2016lookahead} & 91.3 & 92.1 & 91.7 \\
    \newcite{dyer2016rnng} (R) & & & 93.3   & 
    \newcite{sternandreasklein:2017:Long} top-down & 90.4 & 93.2 & 91.8 \\
    \cline{5-8}
    \newcite{kuncoro2017rnng} (R) & & & 93.6 & 
    
    \bf BinarySpan & 91.9 & 92.2 &	92.1 \\ 
    
    \newcite{liu2017inorder} (R) & & & 94.2 & 
    
    \bf MultiSpan & 92.2 &	92.5 & \bf 92.4 \\
    \newcite{fried-stern-klein:2017:Short} (ES) & & & \bf 94.7 & 
    \bf BiaffineRule & 92.0 & 92.6  &	92.3 \\
    \hline
  \end{tabular}
  \caption{Results on the PTB test set.  \emph{S} denotes parsers using auto parsed trees. \emph{E}, \emph{R} and \emph{ST} denote ensembling, reranking and self-training systems, respectively. }
  \label{tab:expeng}
\end{table}

\textbf{Chinese.} Table \ref{tab:expchn} shows the parsing performance on CTB 5.1 test set. Under the same settings, all the three models outperform the state-of-the-art neural model \cite{dyer2016rnng,liu2017inorder}. Compared with the in-order transition-based parser, our best model improves the labeled F1 score by 1.2 ($86.1\rightarrow87.3$).  In addition, \texttt{MultiSpan} and \texttt{BiaffineRule} achieve better performance than the reranking system using recurrent neural network grammars \cite{dyer2016rnng} and methods that do joint POS tagging and parsing \cite{wang2014joint,wang2015feature}.

\begin{table}[!t]
\centering
  \small
  \setlength{\tabcolsep}{3.5pt}
  \begin{tabular}{l|l|l|l|l|l|l|l}
    \hline
    Parser   &    LR & LP &  LF & Parser   &    LR & LP &  LF \\
    \hline
    \newcite{charniak2005rerank} (R) & 80.8 & 83.8 &  82.3 &
     \newcite{petrov2007unlex} & 81.9 & 84.8 & 83.3 \\
     
    \newcite{zhu2013acl} (S) & 84.4 & 86.8 & 85.6 &
    \newcite{zhang2009tran} & 78.6 & 78.0 & 78.3 \\
    \newcite{wang2015feature} (S) & & & 86.6 &
     \newcite{watanabe2015transition} & & & 84.3 \\
    \newcite{huang2009selftraining} (ST) & & & 85.2 & 
     \newcite{dyer2016rnng} & & & 84.6   \\
         \cline{5-8}
    \newcite{dyer2016rnng} (R) & & & 86.9   &
    {\bf BinarySpan}   & 85.9 & 87.1 & 86.5\\

    \newcite{liu2016lookahead} & 85.2 & 85.9  & 85.5 &
    {\bf MultiSpan} & 86.6 & 88.0		& \bf 87.3 \\
    \newcite{liu2017inorder} & & & 86.1  &  {\bf BiaffineRule}&	87.1 & 87.5  & \bf 87.3 \\
    \hline
  \end{tabular}
  \caption{Results on the Chinese Treebank 5.1 test set. }
  \label{tab:expchn}

\end{table}

%% file: part/analysis2.tex
\section{Analysis}
\textbf{Constituent label.} Table \ref{tab:conlabelf1} shows the LF scores for eight major constituent labels on PTB test set.  \texttt{BinarySpan} consistently underperforms to the other two models. The error distribution of \texttt{MultiSpan} and \texttt{BiaffineRule} are different.  For constituent labels including SBAR, WHNP and QP, \texttt{BiaffineRule} is the winner. This is likely because the partition point distribution of these labels are less trivial than other labels.  For NP, PP, ADVP and ADJP, \texttt{MultiSpan} obtains better scores than \texttt{BiaffineRule}, showing the importance of the explicit type information for correctly identifying these labels.  In addition, the three models give similar performances of VP and S, indicating that simple local classifiers might be sufficient enough for these two labels. 

\begin{figure}[!t]
\centering
\begin{minipage}{.45\textwidth}
  \centering
  \includegraphics[width=\linewidth]{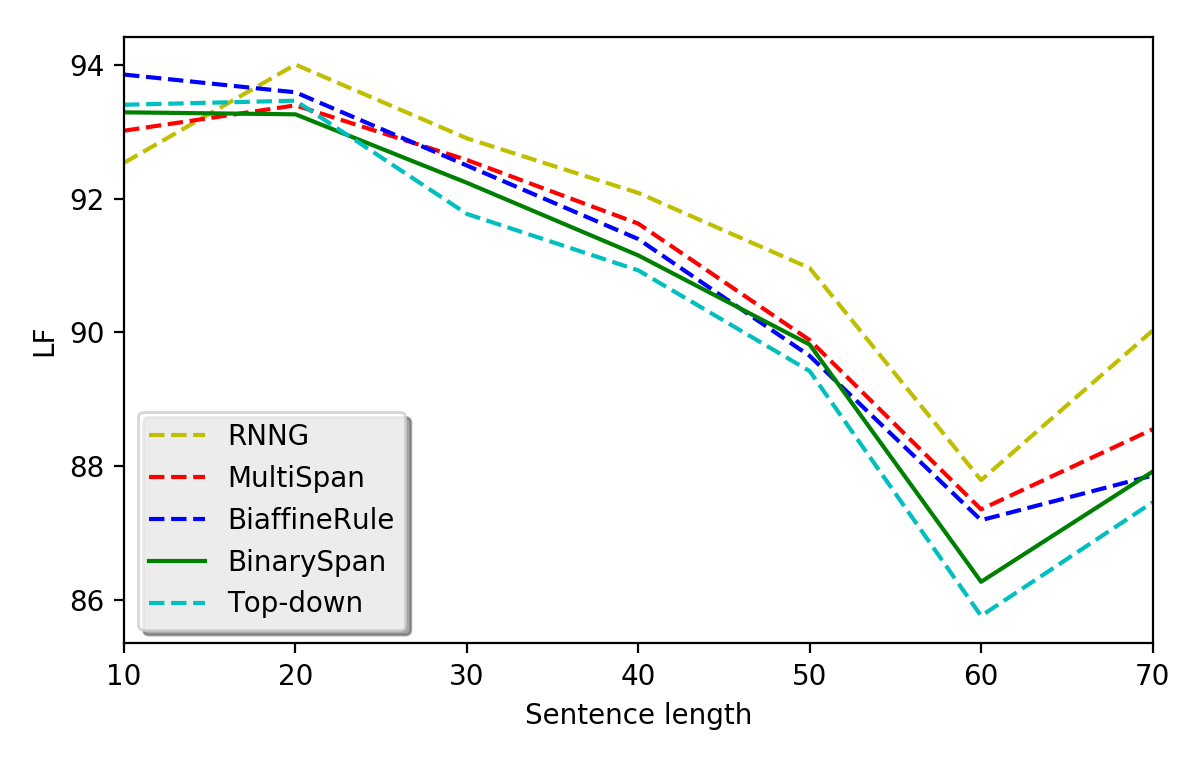}
  \caption{Sentence length v.s LF scores.}
  \label{fig:sentlen}
\end{minipage}
\begin{minipage}{.45\textwidth}
  \centering
  \includegraphics[width=\linewidth]{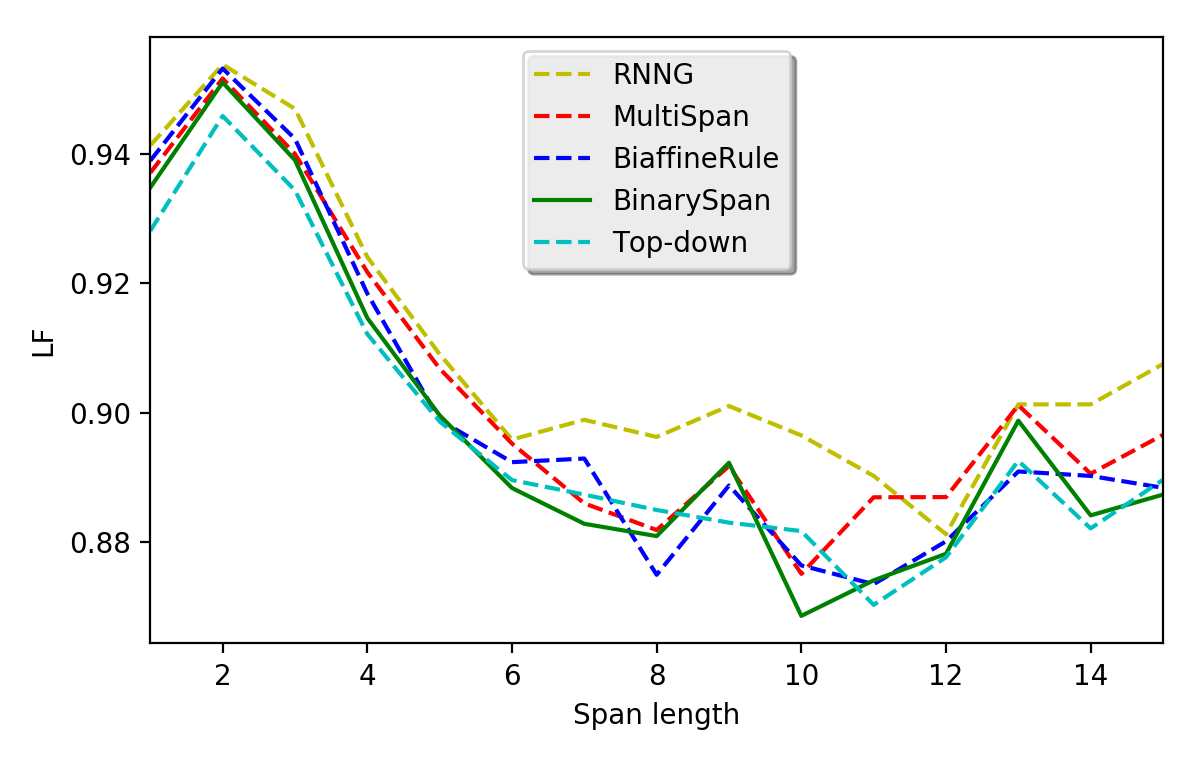}
  \caption{Span length v.s LF scores.}
  \label{fig:spanlen}
\end{minipage}
\end{figure}

\textbf{LF v.s. Length.} Figure \ref{fig:sentlen} and Figure \ref{fig:spanlen} show the LF score distributions against sentence length and span length on the PTB test set, respectively. We also include the output of the previous state-of-the-art  top-down neural parser \cite{sternandreasklein:2017:Long} and the reranking results of transition-based neural generative parser (RNNG) \cite{dyer2016rnng}, which represents  models that can access more global information. For sentence length, the overall trends of the five models are similar. The LF score decreases as the length increases, but there is no salient difference in the downing rate (also true for span length $\leq$6), demonstrating our local models can alleviate  the label bias problem. \texttt{BiaffineRule} outperforms the other three models (except RNNG) when the sentence length less than 30 or the span length less than 4. This suggests that when the length is short, the rule model can easily recognize the partition point.  When the sentence length greater than 30 or the span length greater than 10, \texttt{MultiSpan} becomes the best option (except RNNG), showing that for long spans, the constituent label information are useful. 

\begin{table*}[!t]
    \small
       \setlength{\tabcolsep}{2.5pt}
       \begin{minipage}[b]{.65\linewidth}
         \begin{tabular}{l|c|c|c|c|c|c|c|c|c}
    \hline
        Model & NP & VP & S	& PP & SBAR & ADVP & ADJP & WHNP & QP \\
        \hline 
        BinarySpan & 93.35 & 93.26 & 92.55 & 89.58 & 88.59 & 85.85 & 76.86 & 	95.87 & 89.57 \\
        MultiSpan & \bf 93.61 & 93.41 & 92.76 & \bf 89.96 & 89.16 & \bf 86.39 & \bf 78.21 & 	95.98	& 89.51 \\
        BiaffineRule & 93.53 & \bf 93.46 & \bf 92.78 & 89.30 & \bf 89.56 & 85.89 & 	77.47 & 	\bf 96.66 & \bf 90.31 \\
        \hline 
    \end{tabular}
    \caption{LF scores for major constituent labels.}
    \label{tab:conlabelf1}
    \end{minipage}
    \begin{minipage}[b]{.35\linewidth}
      \begin{tabular}{l|c|c|c}
    \hline
       Model &  UF &  LF  & Speed(sents/s) \\
       \hline 
       BinarySpan & 92.16 & 96.79 & 22.12 \\
       MultiSpan & 92.50 &  97.03 & 21.55\\
       BiaffineRule & 92.22 & 97.12 & 6.00 \\
       \hline 
    \end{tabular}
    \caption{UF, oralce LF and speed. }
    \label{tab:unlabeledf1}
    \end{minipage}
\end{table*}

%% file: part/relatedwork.tex
\section{Related Work}




Globally trained discriminative models have given highly competitive accuracies on graph-based constituent parsing. 
The key  is to explicitly consider connections between output substructures in order to avoid label bias. 
State-of-the-art statistical methods use a single model to score a feature representation for all phrase-structure rules in a parse tree 
\cite{taskar2004maxmargin,finkel2008crf,carreras2008tag}. More sophisticated features that span over more than one rule have been used for reranking \cite{huang2008forest}. \newcite{durrett2015crf} used neural networks to augment manual indicator features for CRF parsing. 
 Structured learning has been used for transition-based constituent parsing also \cite{sagae2005classifier,zhang2009tran,zhang2011syntactic,zhu2013acl}, and neural network models have been used to substitute indicator features for transition-based parsing  \cite{watanabe2015transition,dyer2016rnng,goldberg2014tabular,KiperwasserG16a,cross2016bilstm,coavoux2016tranoracle,shi-huang-lee:2017:EMNLP2017}.

  Compared to the above methods on constituent parsing, our method does not use global structured learning, but instead learns local constituent patterns, relying on a bi-directional LSTM encoder for capturing non-local structural relations in the input. Our work is inspired by the biaffine dependency parser of \newcite{Dozat2016biaffine}. Similar to our work,    \newcite{sternandreasklein:2017:Long} show that a model that bi-partitions spans locally can give high accuracies under a highly-supervised setting. Compared to their model, we build direct local span classification and CFG rule classification models instead of using span labeling and splitting features to learn a margin-based objective. Our results are better although our models are simple. In addition, they collapse unary chains as fixed patterns while we handle them with an encoder-decoder model.  

%% file: part/conclusion.tex
\section{Conclusion}

We investigated two locally trained span-level constituent parsers   using BiLSTM encoders, demonstrating empirically the strength of the local models on learning syntactic structures. On standard evaluation, our models give the best results among existing neural constituent parsers without external parses. 

%% file: part/ack.tex
\section*{Acknowledgement}
Yue Zhang is the corresponding author. We thank all the anonymous reviews for their thoughtful comments and suggestions.